\documentclass[letterpaper]{article} % DO NOT CHANGE THIS
\usepackage{aaai24}  % DO NOT CHANGE THIS
\usepackage{times}  % DO NOT CHANGE THIS
\usepackage{helvet}  % DO NOT CHANGE THIS
\usepackage{courier}  % DO NOT CHANGE THIS
\usepackage[hyphens]{url}  % DO NOT CHANGE THIS
\usepackage{graphicx} % DO NOT CHANGE THIS
\usepackage{natbib}  % DO NOT CHANGE THIS AND DO NOT ADD ANY OPTIONS TO IT
\usepackage{caption} % DO NOT CHANGE THIS AND DO NOT ADD ANY OPTIONS TO IT
\usepackage{algorithm}
\usepackage{algorithmic}
\usepackage{comment}

\usepackage{newfloat}
\usepackage{listings}
\DeclareCaptionStyle{ruled}{labelfont=normalfont,labelsep=colon,strut=off} % DO NOT CHANGE THIS
\floatstyle{ruled}
\newfloat{listing}{tb}{lst}{}
\floatname{listing}{Listing}
\usepackage{times}
\usepackage{epsfig}
\usepackage{graphicx}
\usepackage{amsmath}
\usepackage{amssymb}

\usepackage{booktabs}

\usepackage{xcolor}
\usepackage{soul}

\usepackage{color,soul}

\title{MIMT: Multi-Illuminant Color Constancy via Multi-Task Local Surface and Light Color Learning}
\author{
        Shuwei Li\textsuperscript{1}, Jikai Wang, Michael S. Brown\textsuperscript{2}, Robby T. Tan\textsuperscript{1,3}\\
    \textsuperscript{1} National University of Singapore,
    \textsuperscript{2} York University,
    \textsuperscript{3} Yale-NUS College\\
    {\tt\small shuwei@u.nus.edu, jkwang992@gmail.com, mbrown@eecs.yorku.ca,} {\tt\small robby.tan@nus.edu.sg}
}

\usepackage{bibentry}
\begin{document}

\maketitle

\begin{abstract}
The assumption of a uniform light color distribution is no longer applicable in scenes that have multiple light colors.
Most color constancy methods are designed to deal with a single light color, and thus are erroneous when applied to multiple light colors.
The spatial variability in multiple light colors causes the color constancy problem to be more challenging and requires the extraction of local surface/light information.
Motivated by this, we introduce a multi-task learning method to discount multiple light colors in a single input image.
To have better cues of the local surface/light colors under multiple light color conditions, we design a novel multi-task learning framework.
Our framework includes auxiliary tasks of achromatic-pixel detection and surface-color similarity prediction, providing better cues for local light and surface colors, respectively.
Moreover, to ensure that our model maintains the constancy of surface colors regardless of the variations of light colors, a novel local surface color feature preservation scheme is developed.
We demonstrate that our model achieves {\bf 47.1\%}  improvement (from 4.69 mean angular error to 2.48) compared to a state-of-the-art multi-illuminant color constancy method on a multi-illuminant dataset (LSMI).
\end{abstract}

\section{Introduction}
\label{sec:intro}
\begin{figure}[t!]
	\centering
	\includegraphics[width=1\columnwidth] {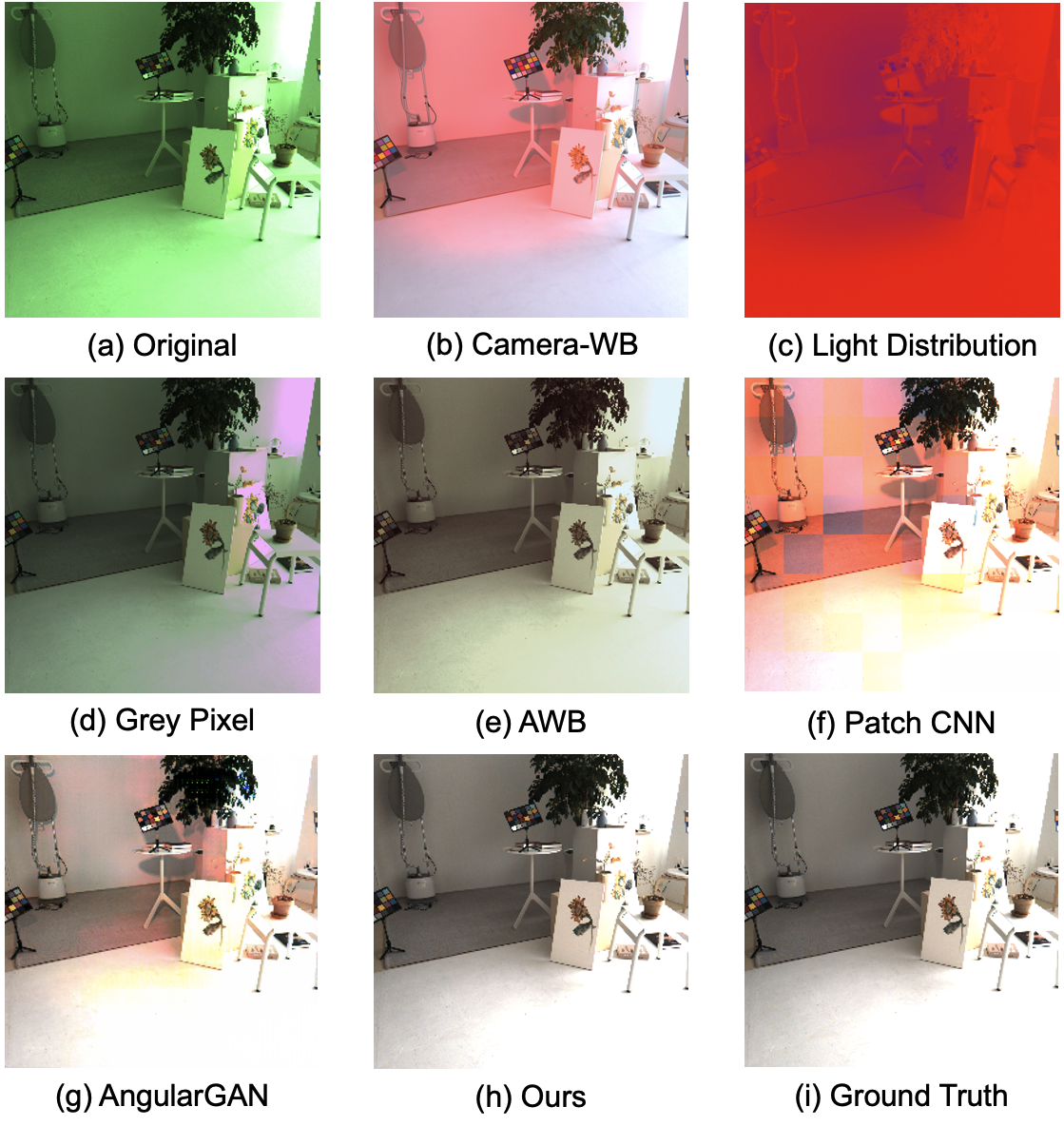}
	\caption{A comparison of our method to state-of-the-art methods targeting a scene with two light colors as the distribution shown in the image (c). The single illuminant color constancy method (b) cannot completely remove the color casts. The illumination color casts are removed from our output image while remaining in other multi-illuminant color constancy results (Grey Pixel~\cite{qian2018revisiting}, AWB~\cite{afifi2022auto}, Patch CNN~\cite{bianco2017single}, and AngularGAN~\cite{sidorov2019conditional}).
	}
	\label{fig:teasurez}
\end{figure}

Computational color constancy determines the dominant color cast caused by the scene's dominant illumination in a sensor image.  Once the sensor's response to the color cast has been estimated, the image can be corrected via a white-balance procedure.  Computational color constancy is one of the key operations applied by cameras and is critical to obtaining an accurate image. 
A number of color constancy methods have been proposed, employing various strategies, e.g., low-level statistical methods,  gamut-based methods, and learning-based methods.  Recently, deep learning has become a mainstream approach due to its impressive performance~\cite{lo2021clcc,bianco2019quasi,xiao2020multi}.
However, most color constancy methods, including state-of-the-art deep learning methods, assume that the scene is illuminated by a single uniform light source.  This assumption is often not valid in many real-world situations, particularly in indoor settings where multiple illuminations are commonly present. When multi-illuminations are present, correcting for only one illuminant can leave strong color casts in the image, as shown in Fig.~\ref{fig:teasurez}.

There has been significantly less work targeting scenes with multiple illuminations. Early work~\cite{gijsenij2011color} applied statistical-based illumination estimation  to image patches.  Similarly, \cite{joze2013exemplar} utilized a nearest-neighbor approach, comparing the statistics of pixels to those in a training set.   These earlier methods were only effective for selected images, highlighting the difficulty of the multi-illumination problem.

Existing deep learning methods tried to solve this problem by performing pixel-wise prediction. Work by \cite{sidorov2019conditional} introduced a GAN-based approach incorporating discriminator loss and a conventional color constancy loss.
While adding the GAN loss improved performance, the method often leaves noticeable spatial artifacts in the corrected image.
Recent work by ~\cite{afifi2022auto} provided a solution that rendered the captured scene to a small set of predefined white-balance settings. The work estimates pixel-wise weights to blend between this set of white-balanced results. 
However, the reliance on a small predefined set of solutions hindered the method's practicality in handling general light colors.

In this paper, we propose a multi-task learning framework that estimates surface color under spatially-varying light colors.
To enhance the ability to capture local light and surface information under multi-light conditions, we design two auxiliary tasks: (1) achromatic-pixel detection and (2) surface color similarity prediction.
Achromatic-pixel detection encourages the primary color constancy module to focus on local achromatic regions that strongly indicate local light colors. Surface color similarity prediction, on the other hand, encourages the model to learn the similarity among pixels of the surface color.
In addition, we employ consistency losses to ensure alignment between outputs of the primary task and the auxiliary tasks. This alignment imposes additional constraints on the primary task's output.
Furthermore,  to ensure that the surface color features are independent from light colors,  we employ the contrastive loss locally on the input image, ground-truth surface colors and predicted surface colors.
Such feature preservation enhances the model's stability of feature extraction when handling varying light colors.

In summary, our contributions are as follows:
\begin{itemize}
	\item We propose an innovative approach to multi-illuminant color constancy by capturing local surface and light information through a multi-task learning framework.
	\item We incorporate two auxiliary tasks into the multi-illuminant color constancy problem, designed to facilitate the acquisition of local light color information and surface color correlations. We devise a new auxiliary task, namely, the surface color similarity prediction task.
	\item To ensure that the extracted local surface-color features are invariant to light color, we propose a feature preservation technique that enforces the input, output, and ground truth images to have the same surface color features, despite their differences in light colors.
\end{itemize}
Our proposed model achieves {\bf 47.1\%} improvement compared to a state-of-the-art multi-illuminant color constancy method on a multi-illuminant dataset (LSMI). In addition, our model maintains a robust performance on the single illuminant dataset (NUS-8) and provides {\bf 39.8\%} improvement on the leading multi-illuminant color constancy method.

\section{Related Work}
Color constancy -- also referred to as white-balancing -- is the process of removing color casts caused by the scene illumination.  The fundamental challenge for color constancy is to estimate the illumination for the image.  The problem can be divided into two scenarios: (1) uniform color constancy, which assumes that there exists only one uniform illumination color in the scene; and (2) multi-illuminant color constancy, assuming more than one illumination color exists.

As mentioned above, the vast majority of methods focus on uniform color constancy. Multi-illuminant color constancy is significantly more challenging because it is necessary to estimate spatially varying illumination color over the image, instead of a single RGB color needed for uniform illumination.  Early works (e.g., \cite{buchsbaum1980spatial, brainard1986analysis, gijsenij2011color}) address both problems using statistic-based features derived from empirical priors.  While these methods required no training data, they had limited ability to handle scenes with multiple illuminations.

Statistic-based methods have been replaced by data-driven approaches using either conventional machine learning, or more recently, neural networks. For example, for uniform color constancy, many CNN-based solutions have been proposed (e.g.,\cite{hu2017fc4, bianco2019quasi, yu2020cascading}. The current state-of-the-art method CLCC by Lo et al~\cite{lo2021clcc} introduces contrastive learning to the uniform color constancy. However, unlike our  local contrastive loss, this method is applied to the whole image globally.

For the multi-illuminant color constancy, the lack of large-scale datasets leads to the few methods in the learning-based approach. For example, Afifi et al.~\cite{afifi2022auto} proposed a method that avoids illumination estimation. Instead, an image is corrected with a predefined set of illuminations.  A CNN is used to estimate per-pixel blending weights among the images.  This approach works well, but only for scenes within the predefined set of illuminations. More relevant to our work is that by Sidorov~\cite{sidorov2019conditional} who proposed AngularGAN, a generative adversarial network (GAN) method trained on a synthetic dataset. This method is considered to be the current state-of-the-art method for multi-illuminant color constancy.  AngularGAN produces good results but often leaves noticeable artifacts throughout the image.   

\begin{figure*}[ht]
	\begin{center}
		\includegraphics[width=2.1\columnwidth] {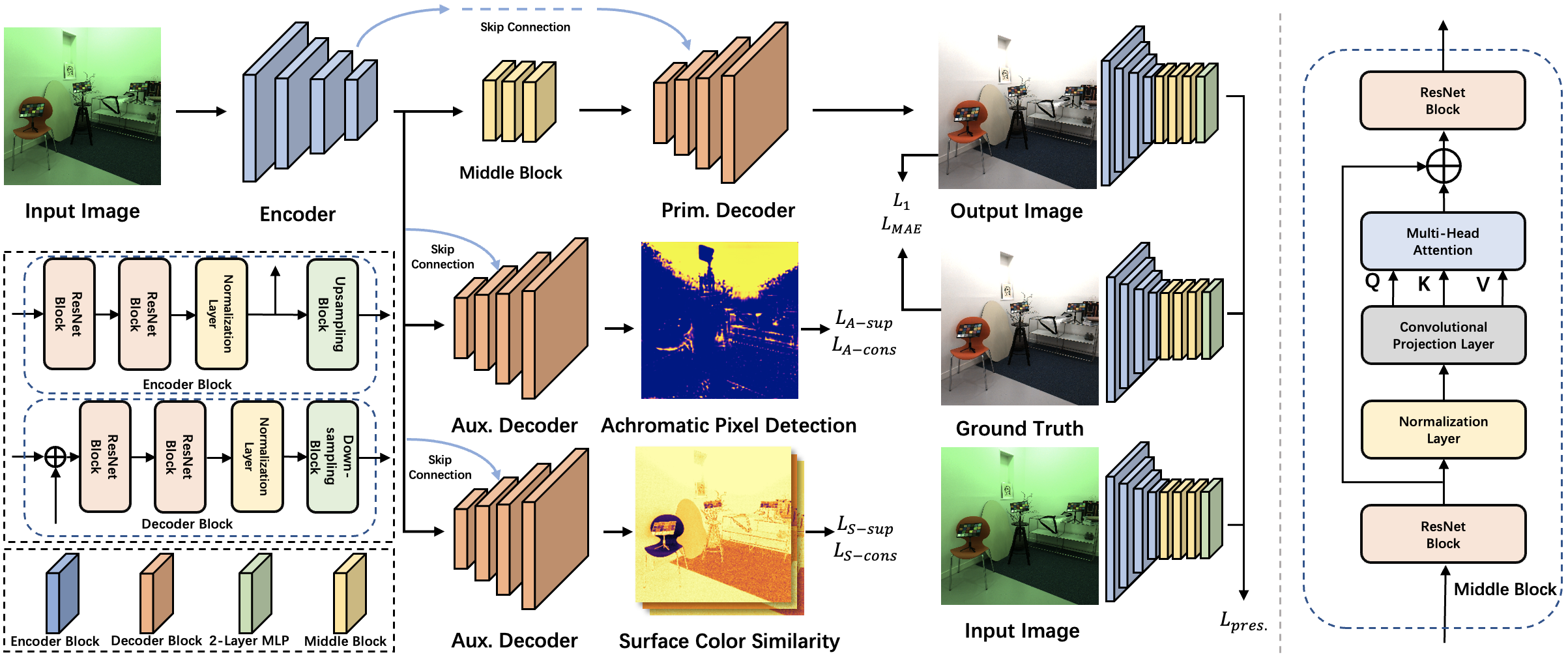}
		\hfill
	\end{center}
	\caption{
		Overview of our network (best viewed in color). Our network has a shared encoder and three individual decoders for each task. Specifically for the primary task, middle blocks are implemented for long-range interactions. The outputs of the three tasks are supervised by the ground truth image. The configurations of the encoder and decoder are shown on the left, while the configuration of the middle block is shown on the right.
	}
	\label{fig:base}
\end{figure*}

\section{Proposed Method}
Given an input image under multiple light colors, our goal is to restore its white-balanced counterpart as if the scene were illuminated by a uniform achromatic light.
The corresponding pixel-wise light color map can be calculated blue from the input image and our output image (i.e., the white-balanced image) and used for error calculation.
As shown in Fig.~\ref{fig:base}, our network utilizes multi-task learning, with color constancy as the primary task, and achromatic-pixel detection and surface-color similarity prediction as auxiliary tasks.
These auxiliary tasks are trained jointly with the primary task and function as constraints.

Within our U-shaped architecture, three branches follow a shared encoder.
Each branch is equipped with its own decoder to carry out specific predictions.
In the training stage, the ground truths of the two auxiliary tasks are calculated from the ground truths of white-balanced images.
For the primary color constancy task, a local surface color feature preservation scheme is exploited as an important constraint.
It should be noted that auxiliary tasks and feature preservation algorithm are necessary only during training, and do not add any computational complexity during testing.

\subsection{Multi-Task Learning Structure}
% Why we use multi-task learning
Color constancy presents an ill-posed problem.
On each pixel, there are two unknowns: illuminant color and surface color.
Furthermore, in multi-illuminant scenarios, as illuminant colors can vary across pixels, learning the mapping between the input image and the output white-balanced image can pose challenges.
Multi-task learning can enhance the model's generalization by acquiring knowledge from interconnected tasks~\cite{zhang2021survey}.
To better tackle the issue of multiple illuminant color constancy, we devise two auxiliary tasks capable of interacting with both light color and surface color.
Specifically, the achromatic-pixel detection task guides the model in learning local illuminant color information inherent in achromatic pixels.
Simultaneously, the surface-color similarity task predicts the similarity between surface colors among pixels.
These two tasks aid in reinforcing the features required for the primary color-constancy task, while also providing consistency losses that ensure the primary task's output aligns with the pertinent information learned through the auxiliary tasks.

\paragraph{Achromatic-Pixel Detection}
Achromatic pixels refer to the object surfaces that have no colors, such as white, gray, or black surfaces.
Detecting achromatic surfaces is an important process in some color constancy methods~\cite{bianco2019quasi, rizzi2002color}.
Even when the scene is lit by multiple light colors, by knowing which surfaces are achromatic, a model can directly estimate the local illuminant colors based on the local input image color.

In the training process, the auxiliary achromatic detection model uses an input image $I$ to predict a weight map, $w$, where $w_{ij}\in[0,1]$ (indexes $i, j$ indicate the pixel location).
The ground truth image $x^\text{gt}$ of the primary color constancy task serves as a reference to determine whether a pixel is achromatic in surface color.
The achromatic pixel detection loss~\cite{bianco2019quasi} for this auxiliary task can be expressed as:
\begin{eqnarray}
		L_\text{A-sup}(x^\text{gt},w) = 1- \frac{\sum_c \sigma_c(x^\text{gt},w)}{\epsilon+\sqrt{3\sum_c \sigma^2_c(x^\text{gt},w))}},
		\label{eq:A}
\end{eqnarray}
where $\epsilon = 10^{-4}$, and  $c \in\{rgb\}$ indicates the color channel. $\sigma_c$ is defined as:
\begin{eqnarray}
	\sigma_c(x,w) = \frac{\sum_{i=1}^{H}\sum_{j=1}^{W}x^c_{ij}w_{ij}}{Z},
\end{eqnarray}
where $Z$ is a normalization factor. $x^c_{ij}$ is the pixel intensity value for $c$ color channel at pixel located at $i,j$.
This loss encourages the weight map to give a higher weight to the achromatic pixels.
The loss only decreases to nearly zero when the three elements in $\sigma$ are closely aligned with each other, signifying an achromatic pixel.

More importantly, the output from the primary color constancy task should align with the learned local light information from this auxiliary task. 
Specifically, when applying the predicted weight map $w$ to the predicted white-balanced image from the primary task, $x^\text{primary}$, the weighted average should be achromatic as well. 
This requirement ensures that the achromatic pixels identified in the auxiliary task match the white balance corrections made in the primary task.
Therefore, the consistency loss is defined as $L_\text{A-cons}(x^\text{primary},w)$ using Eq.~(\ref{eq:A}).

\paragraph{Surface-Color Similarity Prediction}

\begin{figure}
	\begin{center}
		\includegraphics[width=\columnwidth] {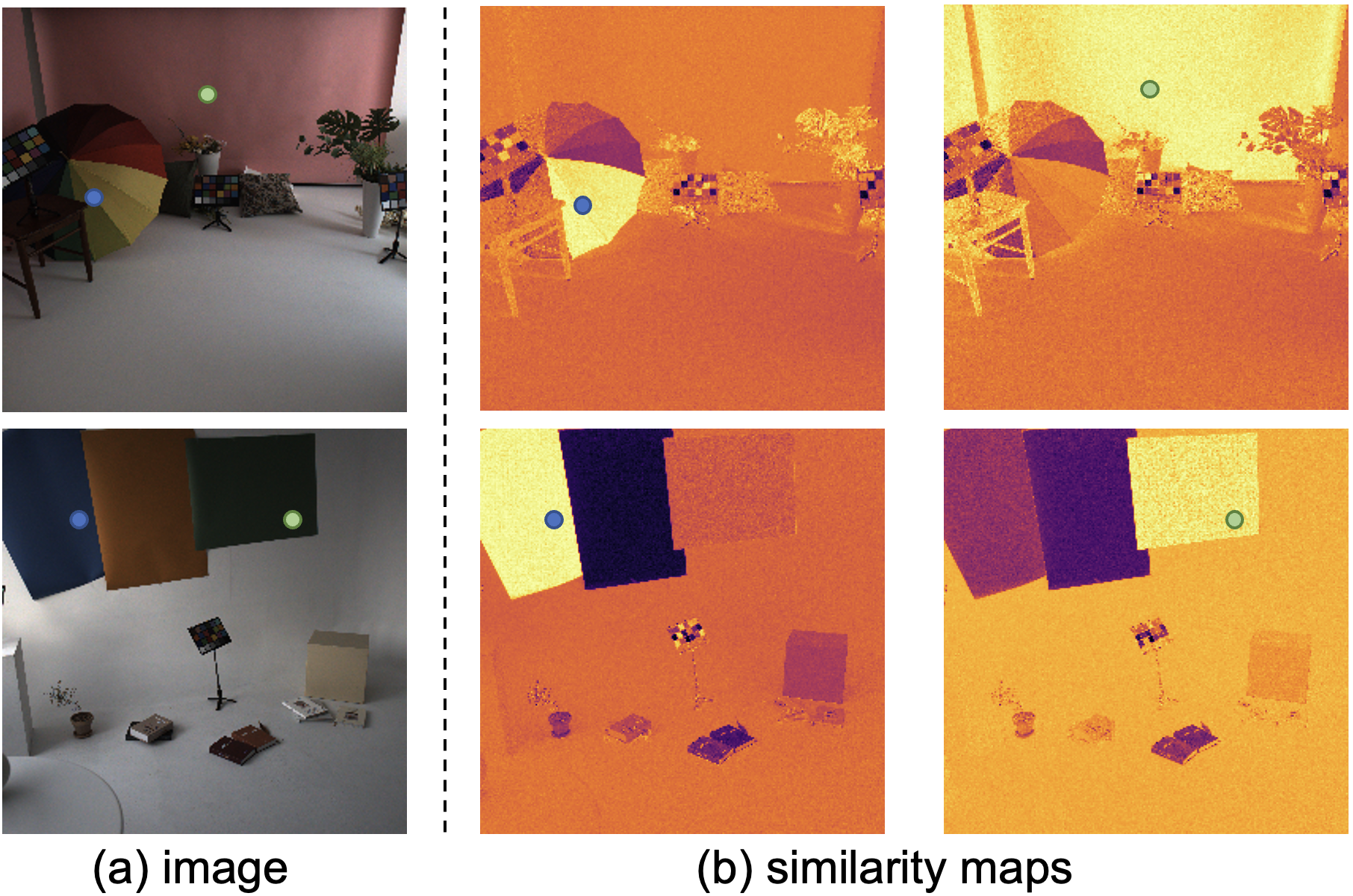}
		%\hfill
	\end{center}
	\caption{
		Ground truth images and their similarity maps. The anchor pixels are indicated in blue and green dots. The brighter areas on the map correspond to regions with a higher similarity in surface color to the anchor pixel (the middle column's anchor pixel is the blue dot). By predicting these similarity maps, our model can better identify areas with both similar and dissimilar colors to the anchor pixel.
	}
	\label{fig:sim}
\end{figure}
Understanding the similarity of the surface colors in the input image is crucial, particularly when dealing with scenes lit by multiple light colors.
By learning the similarity, the model can recognize which areas of the scene have similar (or dissimilar) surface colors, regardless of the variations in light colors that may illuminate those pixels with different colors.
For this reason, we propose surface-color similarity prediction.

During the training process, the ground truth white-balance image $x^\text{gt}$ is divided to $k \times k$ patches (where $k$ can be defined empirically).
We designate the center pixel of each patch as an 'anchor pixel', denoted as  $x^m_\text{anchor}$, where $m$ is the index for the anchor pixels.
We create a ground truth similarity map $S^\text{gt}$ for each anchor pixel.
Each pixel $(i,j)$ in this map is calculated as:
\begin{eqnarray}
	S_{ij}^\text{gt}(x^\text{gt}, x^m_\text{anchor}) = \arccos{\frac{x^\text{gt}_{ij}\cdot x^m_{\text{anchor}}}{\|x^\text{gt}_{ij}\|\|x^m_{\text{anchor}}\| }}.
	\label{eq:sim}
\end{eqnarray}
Hence, if there are $M$ anchors, we will have $M$ similarity maps for the ground truth, $x^\text{gt}$.
This equation calculates the angle between the RGB vectors of the anchor pixel $x^m_\text{anchor}$ and the other pixels $x^\text{gt}_{ij}$ in the image, which provides a measure of their color similarity.

The auxiliary branch in our model predicts a similarity map, denoted as $S^\text{aux-sim}_m$, given the same anchor point, $x^m_\text{anchor}$. Hence,  if there are $M$ anchors, the auxiliary model will predict $M$ similarity maps.
Each of these predicted maps is then compared with our ground truth similarity map $S^\text{gt}_m$, defining our loss:
\begin{eqnarray}
	L_{\text{S-sup}}=\|S^\text{gt}_m-S^\text{aux-sim}_m\|.
\end{eqnarray}
Just as we seek alignment in the achromatic pixel detection using a loss, our predicted surface-color similarity  $S^\text{aux-sim}_m$ should align with the output of our primary task.
Employing the method detailed in Eq.~(\ref{eq:sim}), we compute the similarity map $S^\text{primary}_m$ for the predicted white-balanced image $x^\text{primary}$.
Consequently, the consistency loss between these two similarity maps is given by:
\begin{eqnarray}
	L_{\text{S-cons}}=\|S^\text{primary}_m-S^\text{aux-sim}_m\|.
\end{eqnarray}
The anchor pixels are obtained from the predicted white balanced image from the primary task, $x^\text{primary}$, yet with the same locations as the original $x^m_\text{anchor}$.

\paragraph{U-shaped Structure}
Our network is built upon the transformer and UNet~\cite{ronneberger2015u}.
For each task, the network consists of an encoder, middle blocks (for the primary task), and a decoder.
The encoder is shared by three tasks, while each task has its own decoder.
Our encoder attempts to extract features of local illuminant and surface color correlations.
As Fig.~\ref{fig:base} shows, the middle block consists of two ResNet blocks, a normalization layer, a convolutional projection layer, and a multi-head attention layer.
This transformer-based block can introduce dynamic attention and global context fusion~\cite{wu2021cvt}.

\subsection{Local Surface Color Feature Preservation}
Our primary task is color constancy.
This task focuses on extracting surface-color features, critical for generating a white-balanced image.
Hence, it is important to ensure that these surface-color features remain consistent regardless of the variation in light colors, particularly at the local level.
For this purpose, our idea is to verify whether our middle block outputs (i.e., the surface-color features) for each local patch are consistent when the inputs have different light colors.

\begin{figure}
	\begin{center}
		\includegraphics[width=\columnwidth] {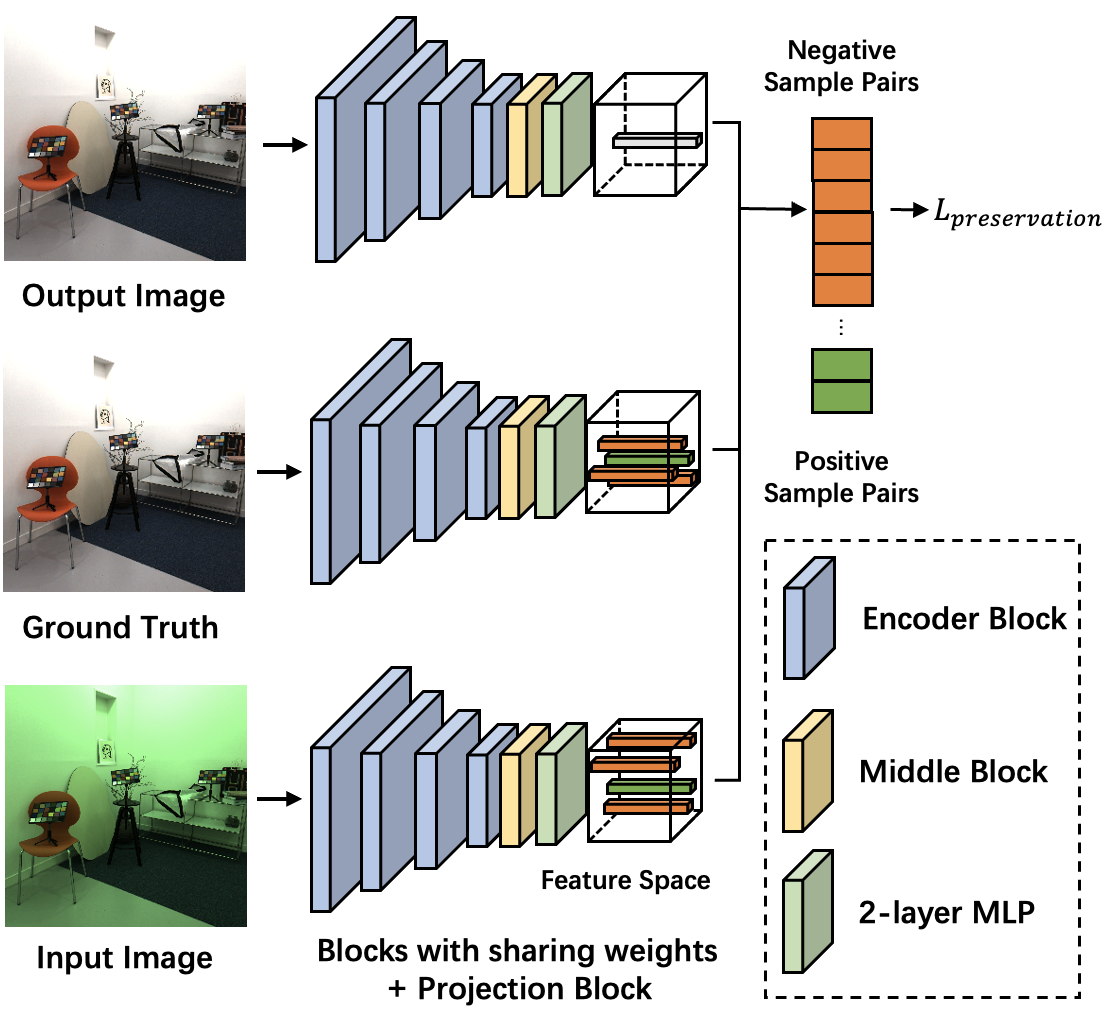}
		%\hfill
	\end{center}
	\caption{
		The structure of local surface color preservation scheme.
		The output image, ground truth, and input image are fed into the encoder and go through middle blocks and a projection block.
		Positive samples (green) have the same location as the anchor sample (gray) on the feature space, while negative samples (orange) have different locations.
	}
	\label{fig:cl}
\end{figure}

Specifically, we employ the contrastive loss~\cite{park2020cut} on the features of our input image, output image (white-balanced image from the primary task), and ground truth image.
We intend to improve the representation learning by pulling positive samples towards the anchor sample while separating negative samples and the anchor sample in the representation space.
As shown in Fig.~\ref{fig:cl}, we use the features in the middle blocks of the primary color constancy task to conduct patch-wise contrastive learning.

The middle blocks' features of the input image, output image, and ground truth are fed into a 2-layer MLP projection block, denoted as $z_i, z_o, z_g$, respectively.
The anchor sample $v^*$ is a vector randomly selected on the feature map $z_o$.
The positive samples $v^+$ are the two vectors on the feature maps $z_i$ and $z_g$ of the same location.
The negative samples $v^-$ are $N$ vectors randomly selected on the feature maps $z_i$ and $z_g$ of different locations.
The decoupled infoNCE (DCE) loss~\cite{yeh2022decoupled} is employed as a contrastive loss, which is based on the cross-entropy loss and calculates the cosine similarity of the samples.
The feature preservation loss can then be formulated as:
\begin{eqnarray}
	%	L_{\text{preservation}} = -\log  \left(\frac{\sum_{i=1}^{2} \exp(v\cdot v^{+}_i / \tau)}{\sum_{i=1}^{2} \exp(v\cdot v^{+} _i/ \tau) + \sum_{i=1}^{N} \exp(v\cdot v^{-}_i / \tau)}\right),
	L_{\text{preservation}} = -\log  \left(\frac{\sum_{i=1}^{2} \exp(v^*\cdot v^{+}_i / \tau)}{ \sum_{i=1}^{N} \exp(v^*\cdot v^{-}_i / \tau)}\right),
\end{eqnarray}
where $N$ is the number of negative samples and $\tau$ is a scaling factor of temperature.
There are two positive pairs and $N$ negative pairs for preservation loss calculation, therefore the loss can be considered as the cross entropy loss of ($N$+2)-way classification.

Note that \cite{lo2021clcc} employs data augmentation on the input images and performs a contrastive loss on the entire image to learn the features for a single light color.
Unlike this method, our approach focuses on local features, which are important in dealing with multiple light colors.
Moreover, our approach applies the contrastive loss locally on the input, output, and ground truth, capturing the same scene but under different lighting conditions.
This enables us to learn the surface color features without any data augmentation.

\paragraph{Total Loss}
Besides the mentioned losses, the $L_1$ loss and the mean angular error (MAE) loss~\cite{sidorov2019conditional} are also used.
To calculate MAE, in brief, the estimated illuminant map is first obtained by dividing the input image by the output image on the pixel level.
Similarly, the ground truth illuminant map is obtained by dividing the input image by the ground truth image.
MAE is the mean of the pixel-wise angular difference between the two maps.

Our total loss is represented as:
\begin{equation}
	\begin{aligned}
		L_{\text{total}} = & \lambda_1 L_{\text{A-sup}}+\lambda_2 L_{\text{A-cons}} + \lambda_3 L_{\text{S-sup}}  + \lambda_4 L_{\text{S-cons}} \\
		& + \lambda_5 L_{\text{preservation}} +\lambda_6 L_1  + \lambda_7 L_{\text{MAE}},
	\end{aligned}
\end{equation}
where $\lambda$'s are the weights of different loss components. $L_1$ is the mean average error loss between the ground truths and predicted images.
%

%%%%%%%%%%%%%%%%%%%%%%%%%%%%%%%%%%%%%%%%%%%%%%%%%%%%%%%%%%%%%%%%%%%%

%------------------------------------------------------------------------
\section{Experiments}
\paragraph{Network Details}
% encoder decoder
In our implementation, the shared encoder consists of 4 encoder blocks while each decoder has 4 decoder blocks.
% middle block
One middle block is implemented between the encoder and decoder in our primary color constancy task.
% contrastive learning details
In our local surface color feature preservation implementation, 16 negative samples are randomly sampled on each image.
The 2-layer MLP has a 512 $\times$ 512 structure.
% patch similarity details
As for surface color similarity prediction, we select $k=3$ in the implementation.

\begin{figure*}[ht!]
	\begin{center}
		\includegraphics[width=2.10\columnwidth] {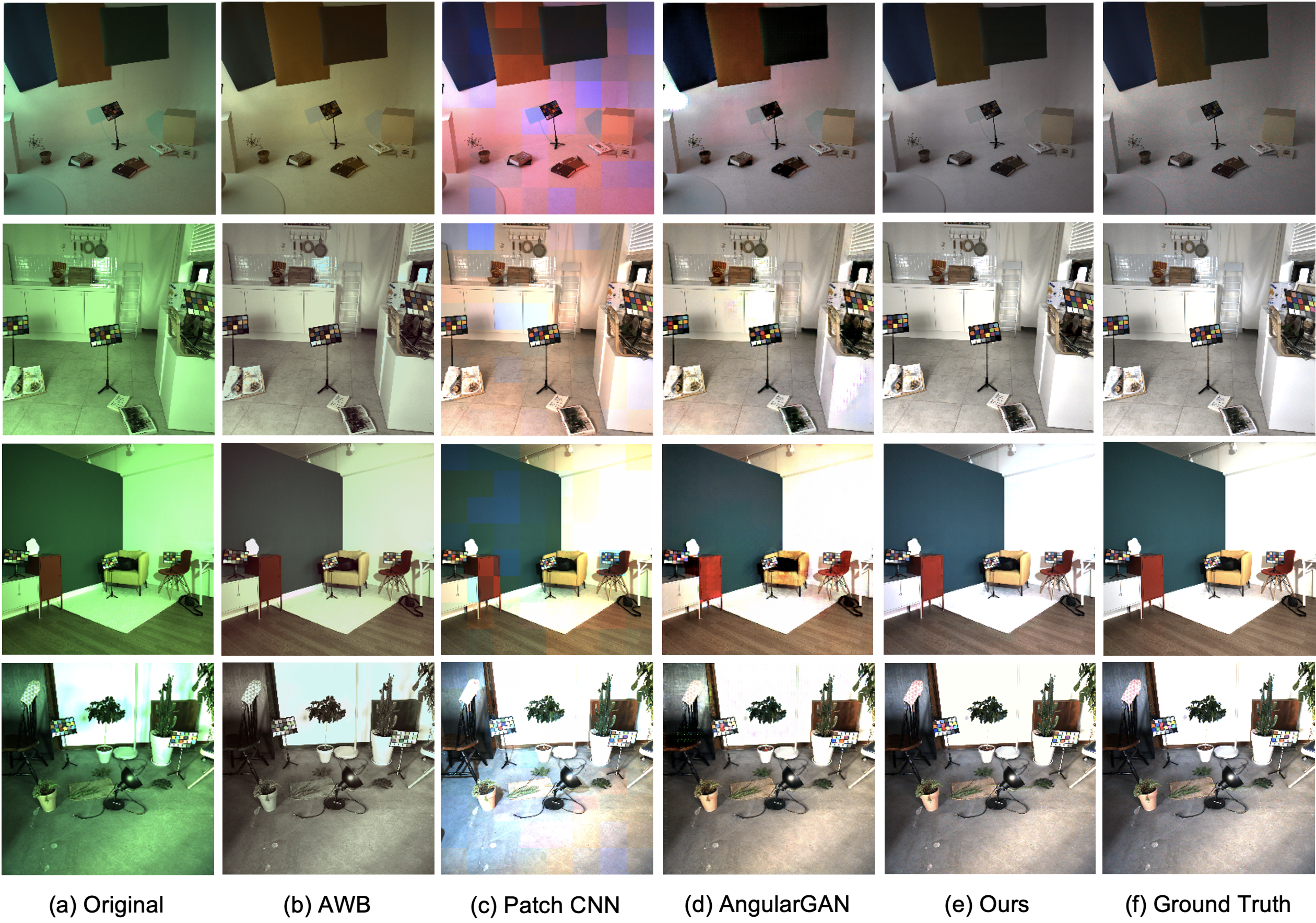}
		\hfill
	\end{center}
	\caption{
		Comparison of the white-balanced images by different multi-illuminant color constancy methods (AWB~\cite{afifi2022auto}, Patch CNN~\cite{bianco2017single}, AngularGAN~\cite{sidorov2019conditional}) on the LSMI dataset~\cite{kim2021large} (best viewed in color).
		The first 3 rows are the comparison in multiple illuminant colors (2 light colors in each image). The 4th row is the comparison in single illuminant color. More samples are in the supplementary material.
	}
	\label{fig:lsmisamples}
\end{figure*}

\paragraph{Training Settings}
% GPU?
The experiments are conducted on a 24GB RTX 3090 GPU.
% image size
The training images are resized to the size of 256$\times$256, while the same resolution is set for the generated images.
% batch size
The batch size is set as 1.
% optimizer
The training of our model uses the Adam~\cite{kingma2014adam} optimizer for 200 epochs using a step decay learning rate scheduler.
The initial learning rate is 0.001 and it decays to 0 after 100 epochs.
The loss weights are all set as 1.

\subsection{Dataset and Evaluation Metrics}
% evaluation sets
We use three color constancy datasets to evaluate our method's performance for multi-illuminant and single-illuminant environments.
To evaluate our method with multiple illumination colors, we employ Large Scale Multi-Illuminant (LSMI) dataset~\cite{kim2021large}.
Both qualitative evaluation and quantitative evaluations are conducted on this dataset.
 To evaluate our method in a single light color setting, we use NUS-8~\cite{cheng2014illuminant} and Cube+ datasets~\cite{banic2017unsupervised}.
During our experiments, we masked all the Macbeth Color Checkers (MCC) by setting the pixel values as (0,0,0) in RGB space.
For training stability, $L_1$ and $L_{\rm MAE}$ loss in our method are not calculated in these regions.

% 1. LSMI dataset
\paragraph{LSMI Dataset}
The large scale multi-illuminant (LSMI) dataset~\cite{kim2021large} contains 7,486 images captured by 3 different cameras on more than 2,700 scenes.
In each scene, there are 1-3 light sources, including natural and man-made light sources.
We use the whole dataset and split it to train, validation, and test sets with the ratio of 0.7, 0.2, and 0.1 randomly.

% 2. NUS dataset
\paragraph{NUS-8 Dataset}
The NUS-8 dataset~\cite{cheng2014illuminant} is widely used in single-illuminant color constancy studies.
It has 1,736 linear raw-RGB images captured by 8 cameras with 210 images for each camera.
%

% 3. Cube+ dataset
\paragraph{Cube+ Dataset}
The Cube+ dataset~\cite{banic2017unsupervised} is commonly used by previous color constancy methods.
In this dataset, 1707 images are collected, including indoor and outdoor images.
\paragraph{Evaluation Metrics}
For multi-illuminant dataset evaluations, we employ the mean angular error (MAE) to evaluate predictions.
MAE measures the pixel-wise angular error between the estimated light color map and ground truth.
Following \cite{akazawa2022n}, we report the performance of the mean and median.
In the evaluation of single-illuminant datasets, we report the angular error in degrees.
To calculate the error, we average the predicted color map and compare it with the ground truth vector. 
We report mean, median, tri-mean of all errors, mean of the lowest and highest 25\%.

\subsection{Multiple Illuminants Evaluation}
\paragraph{Qualitative Evaluation}
% figures to compare
We first conduct visual qualitative comparisons of our method to Patch CNN~\cite{bianco2017single}, AWB~\cite{afifi2022auto}, and AngularGAN~\cite{sidorov2019conditional} on the LSMI~\cite{kim2021large} dataset.
As shown in Fig.~\ref{fig:lsmisamples}, our method generates color-corrected images with higher quality, while the state-of-the-art methods fail to handle complex illuminants.
{Although AngularGAN~\cite{sidorov2019conditional} can have a relatively close white-balancing performance, it often leaves apparent artifacts on the images. In contrast, our results have no such artifacts.
}
Our method also outperforms others on the user study, which is included in the supplementary material.
%
%The experiment setting and results of our user study are in the supplementary material.

\paragraph{Quantitative Evaluation}
\begin{table}[h]
	\caption{Comparison of the overall mean angular error on the LSMI dataset~\cite{kim2021large}. Our method achieves the best result among the methods: Gray-World~\cite{buchsbaum1980spatial}, White-Patch~\cite{land1971lightness}, Gray-Edge~\cite{van2007edge}, PCA~\cite{cheng2014illuminant}, Gray-Pixel~\cite{qian2018revisiting}, Gray-Index~\cite{qian2019cvpr}, Gijsenij~\cite{gijsenij2011color},  Hussain WP~\cite{hussain2018color}, N-WB~\cite{akazawa2022n}, AWB~\cite{afifi2022auto}, Patch CNN~\cite{bianco2017single}, and AngularGAN~\cite{sidorov2019conditional}.}
	\centering
	\footnotesize
	\renewcommand{\arraystretch}{0.6}
	\setlength{\tabcolsep}{6.5mm}{
		\begin{tabular}{lc|c} \toprule
			{Method} & {Mean} & {Median}    \\ \midrule
			{Gray-World} & 11.3 & 8.8   \\
			{White-Patch} & 12.8 & 14.3    \\
			{1st-order Gray-Edge}  & 12.1 & 10.8  \\
			{PCA} & 10.9 & 10.7   \\ \midrule
			{Gray-Pixel(std)} & 16.8 & 17.0 \\ 
			{Gray-Index} & 15.1 & 16.0 \\ 
			{Gijsenij \textit{et al.}} & 18.0 & 17.0   \\
			{Gray-Pixel(M=2)} & 17.1 & 17.4   \\
			{Hussain WP} & 17.7 & 16.9  \\
%			{N-WB (GW)} & 13.9 & 13.1   \\ 
%			{N-WB (WP)} & 9.6 & 8.8   \\ 
%			{N-WB (GE1)} & 12.1 & 10.8   \\ 
			{N-WB (PCA)} & 8.3 & 7.4   \\ \midrule
%			{N-WB (GP(std))} & 12.4 & 11.2   \\  
			{AWB } & 9.54 & 8.19  \\
			{Patch CNN}  & 4.82 & 4.24   \\
			{AngularGAN} & 4.69 & 3.88  \\ \midrule
			{Ours} & \textbf{2.48} & \textbf{2.00}  \\\bottomrule
	\end{tabular}}
	
	\label{tab:multi1}
\end{table}

\begin{table} [h]
	\caption{Comparison of MAE under different number of light colors on the LSMI dataset~\cite{kim2021large}.}
	\centering
	\footnotesize
	\renewcommand{\arraystretch}{1.1}
	\setlength{\tabcolsep}{4mm}{
		\begin{tabular}{l|c|c|c|c} \toprule
			{} &   \multicolumn{2}{c|}{Single} & \multicolumn{2}{c}{Multi} \\ 
			{} & {Mean} & {Median} & {Mean} & {Median}  \\ \hline
			{Ours}  & 2.35 &  1.82 & 2.51 & 2.09 \\\bottomrule
	\end{tabular}}
	\label{tab:multi2}
\end{table}

% compare by the illuminant numbers
We evaluate our method with MAE and also assess it based on its performance under different numbers of light colors.
As shown in Table ~\ref{tab:multi1}, our method outperforms the current state-of-the-art methods in every metric.
To be specific, our method has 47.1\% less mean and 48.5\% less median of the MAE compared to the state-of-the-art method AngularGAN~\cite{sidorov2019conditional}.

Note that, our method shows a robust performance regardless of how many light colors exist in the scene.
{As shown in Table ~\ref{tab:multi2}, our method has a close MAE when facing a multiple number of illuminant colors.}

\subsection{Single Illumination Color Evaluation}

\begin{table}
	\caption{Angular error of various methods on the NUS-8 dataset~\cite{cheng2014illuminant}. By averaging predicted illuminant colors as the final prediction, our method achieves the best results across the metrics.
	}
	\centering
		\footnotesize
	\renewcommand{\arraystretch}{0.8}
	\setlength{\tabcolsep}{1.3mm}{
		\begin{tabular}{lc|c|c|c|c} \toprule
			{Method} & {Mean} & {Median} & {Tri.} & {Best25\%} & {Worst25\%}  \\ \midrule
			{Gray Pixel}  &{13.21}&{13.04}&{12.46}&{3.89}&{20.47}\\
			{Gray Index} &{12.89}&{12.82}&{12.03}&{4.94}&{19.18} \\
			{AWB} &{8.42}&{7.21}&{7.57}&{2.45}&{15.72} \\ 
			{Patch CNN} & {5.51} & {4.52} & {4.86} & {2.95} & {11.51} \\
			{AngularGAN} & {4.10} & {3.75} & {3.89} & {2.16} & {6.47} \\ \midrule
			{Ours}      & \textbf{2.47} & \textbf{1.68} & \textbf{1.82} & \textbf{0.59} & \textbf{5.41} \\\bottomrule
	\end{tabular}}
	\label{tab:NUS}
\end{table}

As shown in Table \ref{tab:NUS}, our method outperforms the state-of-art methods in most of the metrics on the NUS-8~\cite{cheng2014illuminant} dataset, with 39.8\% improvement compared to AngularGAN~\cite{sidorov2019conditional} on the mean value.
Similar results have been achieved on the Cube+ dataset~\cite{banic2017unsupervised}, on which our method outperforms the state-of-the-art multi-illuminant methods.
The result table is included in the supplementary material.
The results demonstrate that our method excels in single illuminant color scenarios, showing robust performance in both single and multiple light color conditions.
	%

%	Besides the average performance in the table, another important fact reflected by the performance metrics in Table~\ref{tab:NUS}, ~\ref{tab:Cube+} is that our method has a more robust performance compared to the previous single/multiple illuminant color constancy methods. For instance, our method provides a 33.6\% improvement on the worst-25\% on the NUS-8~\cite{cheng2014illuminant} dataset and a 48.1\% improvement on the third quartile of $\Delta$E 2000~\cite{sharma2005ciede2000} on the Cube+ dataset~\cite{banic2017unsupervised}.

\subsection{Ablation Studies}

\begin{table}[h]
	\caption{Ablation studies conducted on the LSMI dataset~\cite{kim2021large}. {We compare the mean angular error of our method with and without Achromatic Pixel Detection (APD), Surface Color Similarity Prediction (SCSP), and local surface color feature preservation loss $L_{\text{preservation}}$.}
}
	\centering
	\footnotesize
	\renewcommand{\arraystretch}{0.6}
	\setlength{\tabcolsep}{4.5mm}{
		\begin{tabular}{ccc|cc} \toprule
			{Aux.}&{Aux.}&{Pres.}& {Mean} & {Median}  \\ 
			{APD}&{SCSP}&{Loss}& {} & {}  \\ 
			\midrule
			{}&{\checkmark}&{\checkmark}&{2.99}&{2.33} \\ 
			{\checkmark}&{}&{\checkmark}&{3.05}&{2.56}\\ 
			{\checkmark}&{\checkmark}&{}&{2.84}&{2.19} \\
			{\checkmark}&{\checkmark}&{\checkmark}&{2.48}&{2.00}\\
			\bottomrule
	\end{tabular}}
	\label{tab:as}
\end{table}

\begin{figure}[h!]
	\begin{center}
		\includegraphics[width=1\columnwidth] {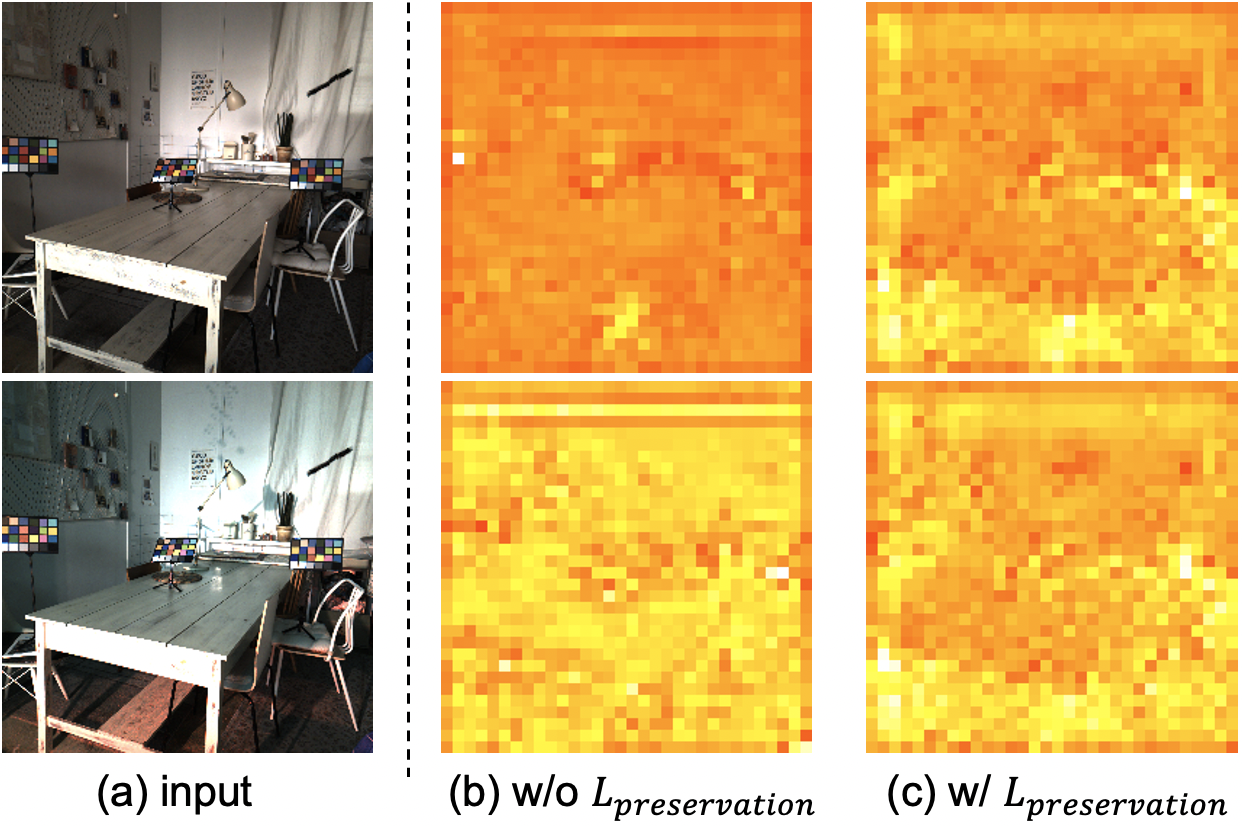}
	\end{center}
	\caption{Comparison of the feature maps. The testing images are taken in the same scene but with different number of light sources. As the illumination changes, the feature maps with $L_{\rm preservation}$ remain virtually unchanged, while those without $L_{\rm preservation}$ change significantly.}
	\label{fig:cl_ab}
\end{figure}

As shown in Table~\ref{tab:as}, we compare the quantitative performance of models trained on different settings on the LSMI dataset \cite{kim2021large}.
By individually removing Achromatic-Pixel Detection (APD), Surface-Color Similarity Prediction (SCSP), and preservation loss $L_{\text{preservation}}$, significant drops in the performance are shown. 
It also shows the best performance when all the proposed components are added.
{Moreover, in Fig.~\ref{fig:cl_ab}, we also compare the feature maps of two images in the testing set in different illuminations with and without our local surface color feature preservation scheme.
	With our proposed scheme, the model can extract stable surface color features when the illuminant conditions are changed.
} 
The ablation studies show that all the components in our proposed method are important.

\section{Conclusion}
	In this paper, to address the multi-illuminant color constancy problem, we proposed a novel multi-task framework.
	To enhance the ability to capture local surface/light information, two auxiliary tasks related to color constancy are designed.
	The achromatic-pixel detection task encourages our model to capture local illuminant color information, while the surface-color similarity prediction task aids in learning surface color correlation. 
	Furthermore, a novel scheme for preserving local surface color features ensures that the extracted features are invariant to changes in illuminant conditions.
	The results of our experiments indicate that our proposed method outperforms existing multi-illuminant approaches and achieves state-of-the-art performance on three different datasets, for both multiple and single illumination corrections.

\bibliography{colorbib}

\end{document}